\documentclass[10pt,twocolumn,letterpaper]{article}

\usepackage{cvpr}              

\usepackage{graphicx}
\usepackage{amsmath}
\usepackage{amssymb}
\usepackage{booktabs}
\usepackage{color, colortbl}
\usepackage{url} 
\definecolor{Gray}{gray}{0.9}

\usepackage[pagebackref,breaklinks,colorlinks]{hyperref}

\usepackage[capitalize]{cleveref}
\crefname{section}{Sec.}{Secs.}
\Crefname{section}{Section}{Sections}
\Crefname{table}{Table}{Tables}
\crefname{table}{Tab.}{Tabs.}

\def\cvprPaperID{*****} 
\def\confName{CVPR}
\def\confYear{2022}

\begin{document}

\title{Video + CLIP Baseline \\ \tt for Ego4D Long-term Action Anticipation}

\author{Technical Report \\ Srijan Das and Michael S. Ryoo\\
Stony Brook University\\
 Stony Brook, NY 11790\\
{\tt\small\{srijan.das, mryoo\}@cs.stonybrook.edu}
}
\maketitle

\begin{abstract}
   In this report, we introduce our adaptation of image-text models for long-term action anticipation. Our Video + CLIP framework makes use of a large-scale pre-trained paired image-text model: CLIP and a video encoder Slowfast network. The CLIP embedding provides fine-grained understanding of objects relevant for an action whereas the slowfast network is responsible for modeling temporal information within a video clip of few frames. We show that the features obtained from both encoders are complementary to each other, thus outperforming the baseline on Ego4D for the task of long-term action anticipation. Our code is available at ~\url{https://github.com/srijandas07/clip_baseline_LTA_Ego4d}.
\end{abstract}

\section{Overview}
\label{sec:intro}
Our goal is to build a framework for long-term action anticipation by leveraging the generalization power of the large-scale pre-trained image-text model CLIP~\cite{CLIP}.
Below we first present the main idea of our approaches in section~\ref{method}. Then, we describe the overall framework used for long-term action anticipation in section~\ref{framework}. Finally, we present our experimental analysis in section~\ref{experiments} and conclude the report in section~\ref{conclusion}.

\section{Our CLIP based Approaches} \label{method}
An activity prediction in Ego4D dataset~\cite{Ego4D2022CVPR} signifies correct prediction of (verb, noun) pair. In order to predict the sequence of activities that will occur in the future from the present visual signal, a fine-grained understanding of the scene is required. Although video models like I3D~\cite{i3d}, and slowfast~\cite{slow_fast} learn spatio-temporal patterns from input stack of frames but is bounded to learn from limited annotated (verb, noun) pairs. 

\begin{figure}
    \includegraphics[width=1\linewidth]{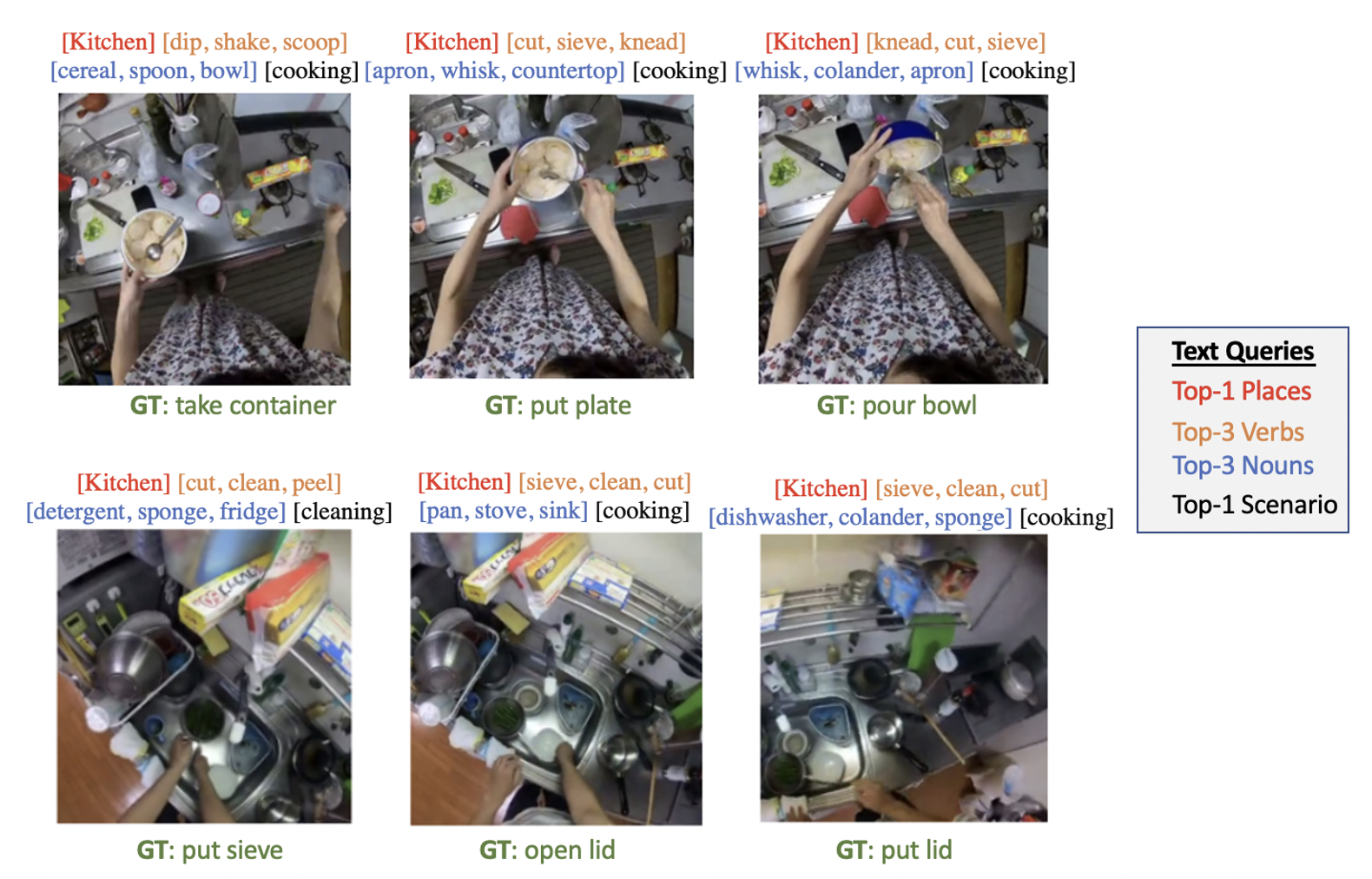} 
   \caption{Qualitative result of CLIP on selected samples of Ego4D dataset. For each sample, we visualize the top-1 place, scenario and top-3 verbs, nouns obtained by computing the similarity between the image and text emebedding. We find that the nouns, places and scenarios realized by the CLIP model is relevant to the ground-truth. However, the verb realized by the CLIP model is extremely noisy due to its limited capability of modeling temporal reasoning.}
\label{qualitative_analysis} 
\end{figure}

On the other hand, large-scale image-text models trained contrastively on paired visual-text data has shown to learn state-of-the-art image representations, especially on zero-shot settings. We confirm in our qualitative analysis (see fig.~\ref{qualitative_analysis}) that the image embedding from CLIP provides a fine-grained understanding of the scene in ego-centric view without the need of any supervision. We also find that the Nouns detected by this model are relevant to the activity performed by the subject but differs with the ground-truth annotations. It also performs well in identifying the place and the scenario of the activity being performed. Thus, we aim at utilizing this large-scale pre-trained model CLIP~\cite{CLIP} for learning salient information in videos for the task of long-term activity prediction. For a unified visual representation of a video $V$, we first extract the image features such that $V = [I_1, I_2, ..., I_N] \in \mathcal{R}^{N\times c}$, via the pre-trained image-text model: CLIP~\cite{CLIP}. Below, we present three variants for visual representation of a sequence of frames based on the extracted CLIP features (illustrated in Fig.~\ref{diag}).

\noindent \textbf{CLIP$_{img}$}: In this variant, we simply aggregate the image representations by performing a mean-pooling across the images to obtain the video descriptor $f_c = \frac{1}{N}\sum_{r=1}^{N}I_r$.

\noindent \textbf{CLIP$_{img}$ + CLIP$_{text}$}: In this variant, we combine the CLIP image embedding (CLIP$_{img}$) and the CLIP text embedding (CLIP$_{text}$) for representing a sequence of frames. The text embedding is obtained by selecting the top-1 Noun, Verb, Scenario, and Place embeddings and concatenating them. The text features are obtained by feeding a prompt to the text backbone of CLIP. The set of verbs, nouns and scenarios are obtained from the ones that occur in Ego4D dataset. The list of places are obtained from~\cite{zhou2017places}.

\noindent \textbf{CLIP attention}: In this variant, we aggregate the sequence of CLIP embeddings across temporal dimension by performing a Multi-headed Attention (MHA) mechanism. We perform a cross-attention between the prompt query and the CLIP image embeddings. The CLIP image embeddings are projected into Keys and Values. The outcome of this MHA is an aggregated video descriptor that performs weighted temporal aggregation based on the learned weights from the text prompt and the image embeddings.

The output of each variant is a video descriptor which is denoted by $f_c$.

\section{Overall Framework} \label{framework}
\begin{figure*}
    \includegraphics[width=1\linewidth]{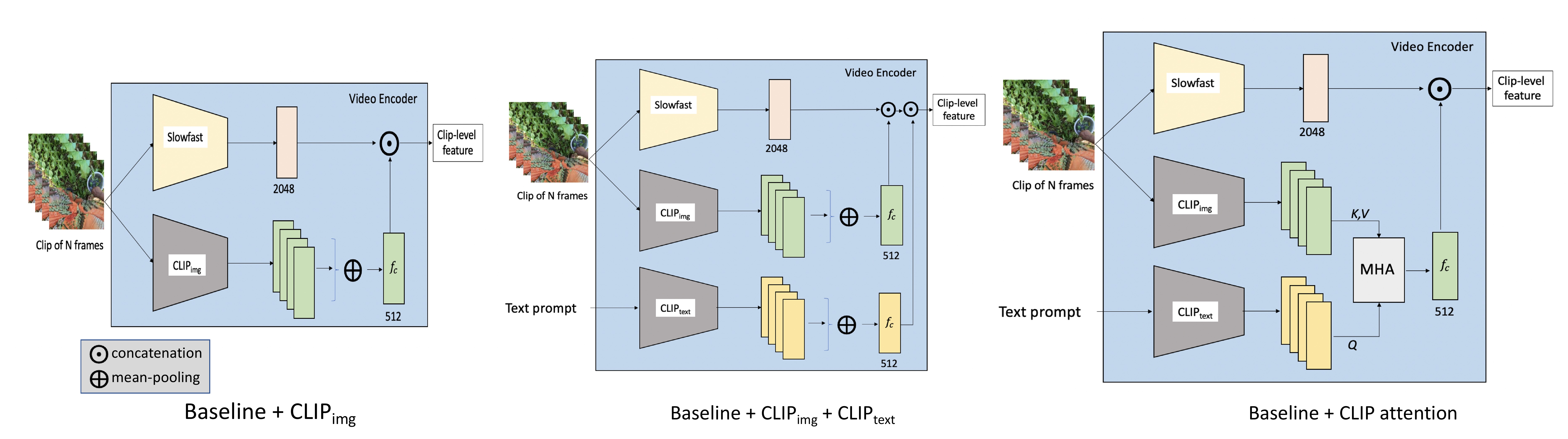} 
   \caption{Video + CLIP baselines for \textbf{Long-Term Action Anticipation} on Ego4D dataset. A clip-level representation is obtained from input N frames using Slowfast, CLIP$_{img}$ and CLIP$_{text}$ encoders. The output is further processed in a transformer aggregation module followed by a decoder network as in~\cite{Ego4D2022CVPR}. MHA denotes Multi-headed Attention in Baseline + CLIP attention framework.}
\label{diag} 
\end{figure*}

The task of predicting future activities given a set of action clips requires modeling of long-term temporal information. Therefore, we adopt the baseline framework provided for long-term action anticipation in~\cite{Ego4D2022CVPR}. The baseline consists of (1) encoder backbone for obtaining clip level features, (2) the aggregation module for combining the obtained features from different clips, and (3) the decoder network for decoding the sequences of future actions. Different from~\cite{Ego4D2022CVPR}, we use two encoder backbones for obtaining clip level features. One is similar to~\cite{Ego4D2022CVPR} - slowfast network that operates on a stack of frames and the other encoder is the CLIP backbone that operates on images. As described in section~\ref{method}, we obtain a video descriptor $f_c$ for an input clip using the CLIP encoder. This video descriptor is concatenated with the video descriptor obtained from slowfast network across channels as illustrated in fig.~\ref{diag}. The rest of the framework remains similar to the baseline~\cite{Ego4D2022CVPR}.

We argue that the CLIP features are aggregated with a mean pooling across temporal dimension which is clearly sub-optimal as explained in~\cite{clip_hitchker}. Thus, we adopt a cross-attention mechanism
between text queries and the image embeddings to assign relevant attention weights to the image embeddings while aggregating into a video descriptor. The resultant features from the complementary encoders are effective for the task of long-term action anticipation compared to the baseline features~\cite{Ego4D2022CVPR}. We confirm this in our experiments. 

\section{Experiments and Analysis} \label{experiments}
\subsection{Training Details}
For training our framework, we follow the exact training recipe provided in~\cite{Ego4D2022CVPR}. For CLIP encoder, we use pre-trained ViT-B/16 without any further fine-tuning. Each short clip consists of 32 frames sampled 4 frames apart. For slowfast encoder, we rely on the pre-trained weights (Kinetics-400 + Ego4D) provided in~\cite{Ego4D2022CVPR}. Thus, our framework training includes training the aggregation module which is a 6 layer transformer with input dimension 2048 + 512 and the decoder network. We train our framework on 8 NVIDIA RTX A5000 GPUs with a batch size of 64 for 30 epochs and a base learning rate of 0.0001. 

\begin{table}[t]
  \centering{%
\caption{Results of the long-term action anticipation task on the val set of Ego4D dataset. Lower is better. The method in Gray is submitted in the Ego4D challenge.}
\begin{tabular}{|l|cc|}
\hline
Method & Verb & Noun \\
\hline
 L1: Baseline~\cite{Ego4D2022CVPR} & 0.745 & 0.779 \\ \hline
 L2: CLIP$_{img}$~\cite{CLIP} & 0.746 & 0.807 \\
 L3: $L_1$ + CLIP$_{img}$ + CLIP$_{text}$ & 0.720 & 0.760 \\
 \rowcolor{Gray} L4: $L_1$ + CLIP$_{img}$ & 0.7151 & 0.748 \\
 L5: $L_1$ + CLIP $attention$ & \textbf{0.7125} & \textbf{0.747} \\
\hline
\end{tabular}\label{results}
}
\end{table}

\subsection{Results}
In Table~\ref{results}, we show our experimental results on the validation set of Ego4D dataset. The results are reported for edit distance at $Z = 20$ and $K = 5$ as in~\cite{Ego4D2022CVPR}. 
The results in L2 clearly shows the demerits of using only CLIP embeddings due to the lack of temporal reasoning of these features.
In L3, we find that the text embeddings do not introduce any complementary features to the CLIP image embedding. 
On the other hand, the CLIP attention framework through its cross-attention proves to be the best strategy to aggregate the CLIP image embeddings in a video clip.  
Consequently, CLIP attention outperforms all the baselines including the Baseline + CLIP$_{img}$ counterpart. Also, the CLIP based video descriptor provides complementary information to the slowfast clip-level representation. Note that L4: Baseline + CLIP$_{img}$ framework is the method which is validated on the test set of Ego4d dataset and secured second position in the Ego4D Challenge for Long-Term Action Anticipation at CVPR 2022.  

\section{Conclusion} \label{conclusion}
In this report, we present a framework that leverages the large-scale pre-trained image-text model: CLIP for the task of long-term action anticipation. 
In our experimental analysis, we find that the CLIP image embedding performs effectively in recognizing Nouns as it requires frame-level understanding. But, these embeddings are not effective in recognizing verbs due to the limitation of the CLIP encoder in modeling temporal information. Thus, pre-trained CLIP, used naively, has some limitations.
Thus, our future work will investigate how to infuse temporal reasoning in the pre-trained image-text models.

\section*{Acknowledgment}
We thank Jinghuan~Shang and Xiang~Li for their technical help in setting up the Ego4D dataset. We thank valuable discussions with members of Robotics Lab at Stony Brook University. 

{\small
\bibliographystyle{ieee_fullname}
\bibliography{egbib}
}

\end{document}